%% file: main.tex
\pdfoutput=1

\documentclass[11pt]{article}

\usepackage{acl}

\usepackage{times}
\usepackage{latexsym}

\usepackage[T1]{fontenc}

\usepackage[utf8]{inputenc}

\usepackage{microtype}

\usepackage{inconsolata}

\usepackage{rotating}
\usepackage{makecell}
\usepackage{multirow}
\usepackage{multicol}
\usepackage{tabularx}
\usepackage{graphicx}
\usepackage{caption}
\usepackage{subcaption}
\usepackage{float}
\usepackage{tcolorbox}
\usepackage[capitalize,nameinlink,noabbrev]{cleveref}
\usepackage{xspace}
\usepackage{todonotes}

\newcommand{\model}{\textsc{AlanaVLM}\xspace}
\newcommand{\dataset}{\textsc{EVUD}\xspace}

\title{\model: A Multimodal Embodied AI Foundation Model for Egocentric Video Understanding}

\author{Alessandro Suglia \textsuperscript{*} 
  Claudio Greco \textsuperscript{*} 
  Katie Baker \textsuperscript{*}  
  Jose L. Part \textsuperscript{*} \\ 
  {\bf Ioannis Papaioannou} 
  {\bf Arash Eshghi}  
  {\bf Ioannis Konstas}
  {\bf Oliver Lemon} \\ 
  Alana AI 
   \\
   \\
    \textsuperscript{*}Equal contribution   
}

\begin{document}
\maketitle
\begin{abstract}
AI personal assistants deployed via robots or wearables require embodied understanding to collaborate with humans effectively. However, current Vision-Language Models (VLMs) primarily focus on third-person view videos, neglecting the richness of egocentric perceptual experience.
To address this gap, we propose three key contributions. First, we introduce the Egocentric Video Understanding Dataset (EVUD) for training VLMs on video captioning and question answering tasks specific to egocentric videos. Second, we present \model, a 7B parameter VLM trained using parameter-efficient methods on EVUD. Finally, we evaluate \model's capabilities on OpenEQA, a challenging benchmark for embodied video question answering. Our model achieves state-of-the-art performance, outperforming open-source models including strong Socratic models using GPT-4 as a planner by 3.6\%. 
Additionally, we outperform Claude 3 and Gemini Pro Vision 1.0 and showcase competitive results compared to Gemini Pro 1.5 and GPT-4V, even surpassing the latter in spatial reasoning. This research paves the way for building efficient VLMs that can be deployed in robots or wearables, leveraging embodied video understanding to collaborate seamlessly with humans in everyday tasks, contributing to the next-generation of Embodied AI\footnote{Code available \url{https://github.com/alanaai/EVUD}}.

\end{abstract}

\input{1_introduction}
\input{2_dataset}

\input{3_training}
\input{4_evaluation}

\input{5_conclusions}
\clearpage
\input{6_limitations}
\clearpage

\bibliography{custom}


\clearpage
\onecolumn
\appendix

\input{appendices/ego4d}
\input{appendices/gemini_prompt}
\input{appendices/ego4d_gemini_data_diversity}
\input{appendices/ego4d_gemini_examples}
\input{appendices/ego4d_gemini_evaluation}
\input{appendices/vsr_prompt}

\input{appendices/egoclip}
\input{appendices/training}
\input{appendices/results}
\input{appendices/alanavlm_evaluation}
\input{appendices/data_model_release}
\end{document}

%% file: 1_introduction.tex
\section{Introduction}
\label{sec:introduction}

Embodied cognition posits that our understanding of the world is fundamentally shaped by our physical bodies and their interaction with the environment~\cite{johnson2015embodied}. Humans leverage this embodied understanding to intuitively grasp physical tasks, anticipate actions, and communicate effectively through nonverbal cues.  For robots and AI systems to become true collaborators, they too must develop a similar understanding.
Egocentric understanding of video data also has key applications in areas such as wearable computing, VR and AR, and video game technology.

\begin{figure}
    \centering
    \includegraphics[width=0.5\textwidth]{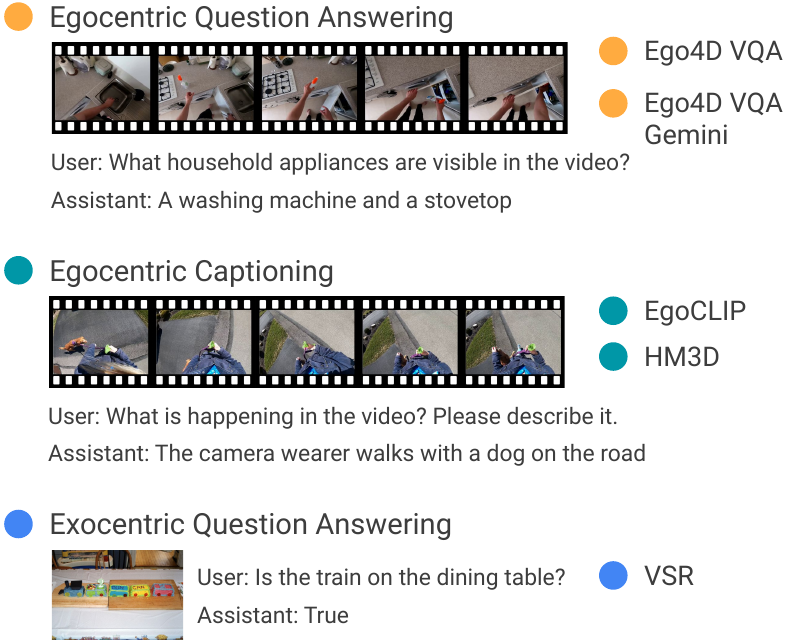}
    \caption{\textbf{Egocentric Video Understanding Dataset (EVUD)}: a collection of egocentric video caption generation and video question-answering tasks that can be used for instruction-tuning video-based VLMs.}
    \label{fig:evud}
\end{figure}

 
In the scenario of an embedded artificial personal assistant, e.g., smart glasses (or a wearable camera for blind and partially sighted people), that can support the user in providing responses to visual queries, we want to build AI systems that can understand videos of the user's activities and of their visual-spatial environment. For this task, it becomes essential that the model is able to receive as input a sequence of frames before generating an answer. 
Recently, by leveraging pre-trained powerful Large Language Models (LLMs), Vision-Language Models (VLMs) have been proposed by using adapters that fuse representations generated by visual experts with textual tokens that can be manipulated by text-only language models (e.g., \citealp{liu2024-llava}). Following this approach, VLMs have been extended to handle video understanding tasks as well (e.g., \citealp{maaz2023-videochatgpt}). However, most of these models have been developed using datasets that include a majority of third-person view videos only ignoring the importance of modelling egocentric videos (e.g., \citealp{xu2017video, caba2015activitynet, maaz2023-videochatgpt}). As demonstrated by \citet{grauman2023ego}, modelling both perspectives is challenging, and dedicated data creation efforts are required to distil this capability into VLMs.

In this paper, we provide a recipe for building VLMs that can solve tasks involving egocentric videos by extending existing video-based VLMs which are trained only on third-person view videos. Concretely, we present three main contributions: 1) we introduce the \textbf{Egocentric Video Understanding Dataset (EVUD)}, a collection of egocentric video caption generation and video question-answering tasks that can be used for instruction-tuning video-based VLMs, which underwent a rigorous human evaluation, 2) we leverage parameter-efficient training to extend existing VLMs and \textbf{train \model} using a limited computational budget; 3) we extensively \textbf{evaluate different model variants on OpenEQA}~\cite{OpenEQA2023}, a challenging real-world benchmark for embodied video question-answering, and achieve state-of-the-art results compared to similarly-sized open-source models and competitive performance with much larger, proprietary variants. We also conducted rigorous human evaluation and quality control of a large portion of EVUD and elicited an error analysis on our system outputs that we hope will inform the next generation of egocentric video-based VLMs.

%% file: 2_dataset.tex
\section{EVUD: Egocentric Video Understanding Dataset}
\label{sec:dataset}

\begin{figure}
    \centering
    \includegraphics[width=0.35\textwidth]{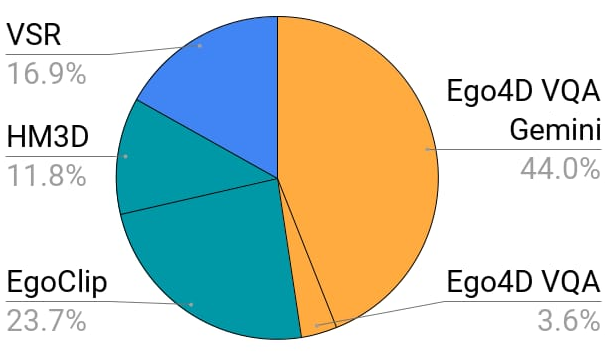}
    \caption{ \dataset is built ensuring  that the majority of examples focus on visual question answering (Ego4D VQA, Ego4D VQA Gemini and VSR), as well as image captioning  (HM3D and EgoClip).}
    \label{fig:evud_distribution}
\end{figure}

We developed the \textbf{Egocentric Video Understanding Dataset (EVUD)} to train VLMs for egocentric video question-answering tasks. This dataset includes 29,477 examples and its components are described below (see \cref{fig:evud_distribution} for an overview). 

\subsection{Ego4D VQA}\label{sec:ego4d_vqa}

We consider the Ego4D collection as a high-quality source of egocentric videos that were collected in diverse settings with different types of cameras~\cite{Ego4D}. Specifically, from the Ego4D NLQ training set, we gathered 13,849 annotated clips extracted from 933 videos (see \cref{sec:AppendixEgo4DPreprocessing}). 
Then, we filtered questions having corresponding human-annotated answers which resulted in 1,137 question-answer (QA) pairs, all of which were added to \dataset.

\subsection{Ego4D VQA Gemini}\label{sec:ego4d_vqa_gemini}

Inspired by LLM-based approaches for generating training data (e.g., \citealp{li2023mimic}; \citealp{Self-Instruct}) and state-of-the-art performance of large multimodal language models, we prompted Gemini Pro 1.5 \citep{Gemini} in a zero-shot multimodal fashion to produce a dataset consisting of 96K question and answer pairs requiring video understanding. These QA pairs belong to seven different categories corresponding to those specified in the OpenEQA episodic-memory question answering dataset \citep{OpenEQA2023}: object recognition, attribute recognition, object state recognition, object localization, spatial reasoning, functional reasoning, and world knowledge.

We gathered all 13,849 clips from Ego4D NLQ (see Section~\ref{sec:ego4d_vqa}) and used them as reference for the following data generation process. Each extracted clip was queried with Gemini Pro 1.5 using the VertexAI API~\footnote{https://cloud.google.com/vertex-ai} using zero-shot multimodal prompting with default settings (see \cref{sec:AppendixGeminiPrompt} for the prompt definition). 

In total, 13,789 of the clips successfully passed the Gemini Pro 1.5 filters, corresponding to 99.6\% of the input clips. Of these clips, 100\% of the outputs were successfully parsed to extract the seven \textit{(category, question, answer)} tuples, resulting in an overall dataset of 96,523 egocentric video QA pairs (see \cref{sec:AppendixEgo4DVQAGeminiDiversity} for a summary and \cref{sec:AppendixEgo4DVQAGeminiExamples} for examples of generated data). To use this data for training, the QA pairs were formatted into a series of QA turns.
In \dataset, we used 12,978 clips between 2 and 60 seconds in length (corresponding to 90,846 QA dialogues).

\subsubsection{Ego4D VQA Gemini Dataset Evaluation}

To evaluate the quality of the generated data, we took a random set of 200 clips (corresponding to 1,400 examples) and one of the authors determined whether the questions, categories, and answers were relevant and correct, following the human evaluation schema of the Self-Instruct dataset (\citealp{Self-Instruct}; details in \cref{sec:AppendixGeminiEvaluation}).

Gemini demonstrated a strong ability to generate appropriate questions tailored to the specified categories and visual context, achieving an overall rate of 87.1\% for appropriate questions and 95.6\% for appropriate categories. However, it performed considerably worse (58.9\%) in generating correct and acceptable answers.
Additionally, the model's proficiency varied across categories, especially with regards to answer correctness (see \cref{fig:human_eval_cqa} in \cref{sec:AppendixGeminiEvaluation}). For object localization, spatial reasoning, and object recognition, fewer than 50\% of the answers were deemed correct and acceptable.





In cases where the model-generated answer was found to be incorrect and/or unacceptable, the gold standard answer was also annotated. These gold standard answers were integrated into \dataset by replacing the model-generated answers for those questions. In this way, 575 examples were updated to human gold annotated answers and 825 model-generated were found to be satisfactory.





\subsection{VSR}

In order to distil fine-grained visual understanding skills into \model, we use the Visual Spatial Reasoning (VSR) dataset~\cite{liu2023visual} as a source of data for generating polar VQA pairs. In particular, for each example in the training set, we give the statement to a language model (\mbox{\textit{Llama-3 8B}}, ~\citealp{llama3modelcard}), and prompt it~\footnote{We use Llama-3 via Ollama and report the prompt we used in~\cref{sec:AppendixVsrPrompt}.} to transform the statement into the corresponding question. Then, we use the truth value associated with the statement to generate an answer, randomly selecting ``True" or ``Yes" for positive answers, or ``False" or ``No" for negative answers. This results in 7,680 examples that are part of \dataset.  
\subsection{EgoClip Captioning}
To further improve \model's visual grounding ability, we also included  
a portion of the 3.8M EgoClip video-caption pairs~\cite{lin2022egocentric}. 
To build our captioning dataset, we sample only clips whose length is between 2 and 60 seconds resulting in 7,000 clips. We then convert the abstracted language in the original captions into natural language prompts using rules (see \cref{sec:AppendixEgoclipPreprocessing}). We used all 7,000 clips with associated captions in \dataset.

\subsection{HM3D Captioning}
The OpenEQA benchmark is composed of two different settings: ScanNet scenes which are very photorealistic~\cite{scannet}, and HM3D scenes which contain many visual artefacts~\cite{hm3d}. Considering that most video-based VLMs are trained on videos recorded in real-world settings, there is a mismatch with HM3D videos. Therefore, inspired by \citet{ehsani2023imitating}, we use the Habitat simulator~\cite{savva2019habitat} to generate the shortest paths to specific objects relevant to the OpenEQA benchmark. Specifically, we first extract all the noun phrases from the OpenEQA benchmark using spaCy \citep{spaCy} to get our candidate set of objects $\mathcal{O}$. Then, for each \textit{training} scene in HM3D, we spawn the agent in a random location and create the shortest paths to all the objects in the current scene which are also in  $\mathcal{O}$. Given these shortest paths, we create 3,475 short videos with associated captions generated using a fixed set of prompts similar to EgoClip (examples in \cref{sec:AppendixEgoclipPreprocessing}) and used them all in \dataset.

%% file: 3_training.tex
\section{Model Training}
We build \model by fine-tuning Chat-UniVi \cite{jin2023chat} --- a vision \& language foundation model equipped with video understanding capabilities --- on \dataset. This fine-tuning step is essential for injecting the \textit{egocentric video understanding} skills that are unique to \model. We decide to build our model starting from Chat-UniVi for several reasons. First, it is an open-source model whose code and weights are publicly available. 
Second, it is designed for handling language, images, and videos 
taking an arbitrary number of frames into account. Third, it outperforms other open-source vision and language foundation models in classic video understanding tasks. 

As follows, we describe the fine-tuning recipe that we used to build our model trying to preserve the original capabilities that were distilled during the instruction tuning stage.
We mitigate the forgetting of previously learned skills by leveraging rehearsal \cite{robins1995catastrophic}, which consists in the retraining of the model on a small percentage of the previously learned information as the model is trained on new information.
We fine-tune our model using Low-Rank Adaptation (LoRa; \citealp{hu2021lora}), which freezes the pre-trained model weights and injects trainable rank decomposition matrices into each layer of the Transformer architecture, greatly reducing the number of trainable parameters for downstream tasks.
In all our experiments, we fine-tune \model on rehearsal data and \dataset. See \cref{sec:AppendixTrainingDetails} for training details.

%% file: 4_evaluation.tex
\section{Evaluation \& Results}
\label{sec:evaluation}









For our evaluation, we use the episodic memory use case of the OpenEQA benchmark~\cite{OpenEQA2023}. OpenEQA uses GPT-4 to rank the appropriateness of the generated answers concerning the ground-truth answers. To favour reproducibility, we use the highly capable open-weight model LLama-3 70B~\cite{llama3modelcard}. 

\begin{table}[!ht]
\centering
\small
\begin{tabular}{|l|c|c|c|}
\hline
\textbf{Model}               & \multicolumn{1}{l|}{\textbf{SN}} & \multicolumn{1}{l|}{\textbf{HM3D}} & \multicolumn{1}{l|}{\textbf{All}} \\ \hline
GPT-4 (text-only)* &             32.5                       &         35.5 & 33.5                                  \\ 
GPT-4V (50f)*           & 57.4                                  & 51.3                               & 55.3                                  \\ 
Claude 3 (20f)*         & n/a                                   & n/a                                & 36.3                                  \\ 
Gemini 1.0 Pro V. (15f)* & n/a                                   & n/a                                & 44.9                                  \\ 
Gemini 1.5 Flash (50f)      & \textbf{74.0}                 & \textbf{69.7}{}              & \textbf{72.5}                 \\ 
Gemini 1.5 Pro (50f)        & 66.9                 & 61.0              & 64.9                 \\ \hline
Chat-UniVi (text-only)            & 43.4                                  & 32.4                               & 39.7                                  \\ 
Chat-UniVi (50f)            & 43.4                                  & 40.4                               & 42.3                                  \\ 
AlanaVLM (50f)              & \textbf{47.8}                                  & \textbf{44.8}                               & \textbf{46.7}                                  \\ \hline
\end{tabular}
\caption{Results on OpenEQA comparing AlanaVLM against other VLMs (with \textit{n}f indicating the number of frames) on ScanNet (SN), HM3D, and all instances. (*): Results taken from \citet{OpenEQA2023}.}
\label{table:results}
\end{table}

We compare several ablations to derive AlanaVLM's best configuration and we report additional details
in \cref{sec:AppendixResults} and \cref{table:AppendixDetailedResults}. In this section, AlanaVLM is the best-performing model which is trained on Ego4D VQA, Ego4D VQA Gemini, VSR, and EgoClip. \cref{table:results} shows the overall performance of AlanaVLM with respect to other VLMs on ScanNet (SN), HM3D, and all OpenEQA instances (All). AlanaVLM outperforms the base model Chat-UniVi by 4.4\%. Despite having only 7B parameters and being fine-tuned with LoRa, AlanaVLM outperforms Gemini 1.0 Pro Vision and Claude 3 and its performance is comparable with all other larger VLMs except GPT-4V and the Gemini 1.5 models. However, AlanaVLM outperforms GPT-4V on spatial questions (\cref{table:AppendixDetailedResults}). We do not attempt full fine-tuning to provide a more cost-effective solution; it is reasonable to expect even better results after this stage is completed (cf. Table E in \citealt{jin2023chat}).

We notice that most models perform better on SN than on HM3D, probably because of its higher quality. Adding HM3D to the \model training doesn't help either, maybe because its descriptions are not fine-grained enough. 
Finally, we note that the most recent version of Gemini surpasses all other frontier models in this task presumably due to its ability to encode higher-resolution video frames leveraging its 1M context length.

\begin{figure}
    \centering
    \includegraphics[width=0.5\textwidth]{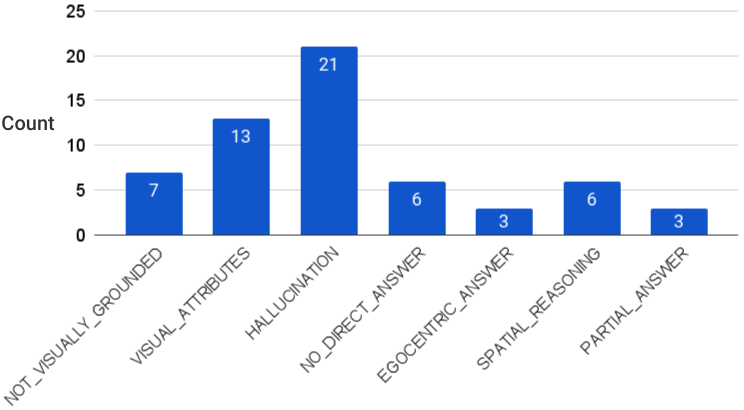}
    \caption{Human error analysis performed on 98 QA pairs on OpenEQA.}
    \label{fig:error_analysis}
\end{figure}

\paragraph{Error analysis} To gain further insights into AlanaVLM predictions, we perform an error analysis based on 98 QA pairs and derive a categorization of the errors. We find that 60\% of answers are incorrect. We notice that the Pearson correlation between human and LLM ratings was $0.76$. Moreover, in roughly 7\% of cases, the LLM said that the answers were wrong even though humans noticed that both answers applied. As shown in \cref{fig:error_analysis}, most errors concern visual attributes (e.g., object colours) or hallucinations (e.g., missing objects). Additionally, we find that \model struggles with spatial reasoning which is required to understand the relationships between objects (error category = \texttt{SPATIAL\_REASONING}). We note that in a few cases \model generates answers that are not aligned with the camera wearer's egocentric point of view (error category = \texttt{EGOCENTRIC\_ANSWER}). This highlights the need for more robust visual encoders for VLMs that can capture fine-grained details of the visual scenes when trained with egocentric vision perception~\cite{pantazopoulos2023multitask}. Finally, we also highlight the problem of current VLMs being overpowered by the original LLM probability distribution which produces not only hallucinations but answers that are not visually grounded or which indirectly answer the question~\cite{guan2024hallusionbench}.

%% file: 6_limitations.tex
\section*{Limitations}

In this paper, we present a training recipe for designing and training VLMs that can perform visual question answering in an embodied setting specifically when receiving a video stream. When designing our training recipe, we made sure that fundamental tasks such as captioning and question answering are well represented in our dataset mixture because they somehow elicit different visual grounding capabilities. To the best of our knowledge, this is the first paper that describes a training recipe for building VLMs able to \emph{generate} responses about egocentric videos. 

Despite its strengths, this paper has some limitations that we acknowledge in this section: 1) \model is trained using LoRa therefore it is not fully leveraging the training on \dataset to the full extent as demonstrated by \citet{jin2023chat}; 2) to avoid potential overfitting and to facilitate fast training times, \dataset includes roughly 39K instances; this is somehow unconventional compared to current training regimes involving millions of examples. However, we don't consider this as a downside of our training recipe because most of the generated datasets in our mixture can be easily scaled up allowing one to further boost performance; and 3) as shown by our quality control evaluation, the Ego4D VQA Gemini data had an accuracy of $58.9\%$ for the generated answers. Relying on frontier models to generate training data inherently has a disadvantage in that the generated training data is only as good as the capability of those models. We ameliorated a small batch via our human control step, but in future, advances in frontier models (e.g. the performance improvement we saw in Gemini 1.5 Flash) may result in more robust vision-language training datasets. 

Finally, it is important to note that, despite its competitive performance on this benchmark, \model still has several important limitations in terms of its visual understanding capabilities based on the careful human error analysis that we performed. Particularly, most of the errors can be considered as visual hallucinations of objects that either are not present in the scene or that are more prominent than the target object. Additionally, more research is required to understand how to design visual resamplers that are able to generate more fine-grained visual representations for the LLM which do not discard important visual attributes and spatial information---another major bottleneck for \model as well as proprietary models such as GPT-4V, and in general of many current VLMs as demonstrated by \citet{pantazopoulos2024lost}. 

\section*{Ethics Statement}

Egocentric video understanding with VLMs presents a powerful new approach to analyzing first-person videos. However, this capability raises significant ethical considerations that must be addressed. 

It is important to prioritize user privacy by ensuring informed consent is obtained for all video data collection. All our datasets are derived from academic benchmarks in which anonymization techniques are employed wherever possible to minimize the risk of identifying individuals within the videos. For instance, we have used Ego4D which has strict policies about the usage of such data.

Another important consideration is the potential for bias in VLM development, particularly if trained on imbalanced datasets. When building \dataset, we made sure to cover diverse and representative datasets during training including both image, first-person videos, and third-person videos. However, we acknowledge that this has to be proven improved when considering the deployment of this \model in the real world. For instance, in household settings like OpenEQA, it is important to make sure that the model is trained on culturally relevant objects without favouring western-centric object distributions~\cite{liu2021visually}. 

%% file: appendices/ego4d.tex
\section{Ego4D Preprocessing}
\label{sec:AppendixEgo4DPreprocessing}

\begin{figure}[htbp]
\centering
\includegraphics[width=0.8\linewidth]{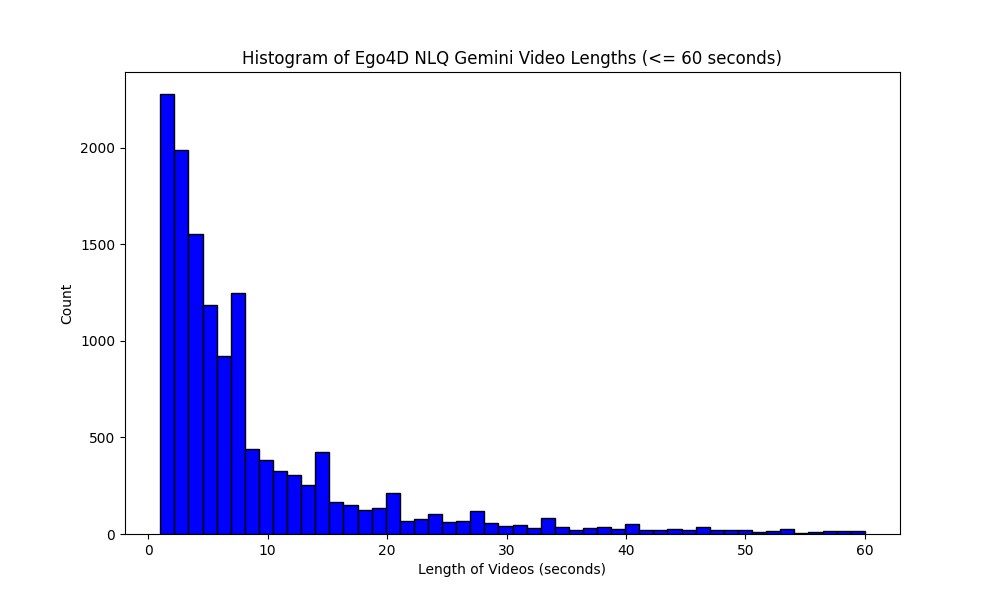}
\caption{Length distribution of Ego4D NLQ clips.}
\label{fig:ego4d_gen_clip_lengths}
\end{figure}

We designed a preprocessing step to utilize the Ego4D NLQ videos. Specifically, 13,849 Ego4D NLQ clips were extracted by slicing the original 933 NLQ training set videos from $int(min(0, clip\_start))$ to $int(max(video\_length, clip\_end))$ for each clip \citep{Ego4D}. The mean clip length is 12.1 seconds, with a min of 1.0 seconds and max of 481.0 seconds. See \cref{fig:ego4d_gen_clip_lengths} for a distribution of the lengths for the 13,355 clips of $\leq60$ seconds length.

%% file: appendices/gemini_prompt.tex
\section{Gemini Pro 1.5 Prompt}
\label{sec:AppendixGeminiPrompt}

To generate the Ego4D VQA Gemini dataset we adapted the prompt from MIMIC-IT. We updated it to the video modality and to obtain a question and answer pair for each of the seven OpenEQA task categories (\citealp{li2023mimic}; \citealp{OpenEQA2023}):

\begin{tcolorbox}[width=\textwidth, colback=gray!10, colframe=black!100, title=Ego4D VQA Gemini prompt]
\texttt{[INPUT VIDEO]} \\

You are an intelligent embodied agent that can answer questions. You will be shown a video that was collected from a single location. \\

Your task is to generate a question for each of the following categories: object recognition, attribute recognition, object state recognition, object localisation, spatial reasoning, functional reasoning, world knowledge. \\

Ask diverse questions and give corresponding short answers. Include questions asking about the visual content of the video. The questions you posed can include the actions and behaviors of people or objects in the video, the chronological order of events, and causal relationships. Only include questions that have definite answers. Do not ask any questions that cannot be answered confidently. \\

Don't use headers. You should use the following format for each category: \\

Category: <category>

Question: <question>

Short answer: <answer> \\

Assistant:
\end{tcolorbox}

%% file: appendices/ego4d_gemini_data_diversity.tex
\section{Ego4D VQA Gemini Dataset Diversity}
\label{sec:AppendixEgo4DVQAGeminiDiversity}

Summary of question and QA pair counts generated by Gemini are shown in \cref{tab:qanda} for each category. Overall, a diverse set of questions were generated with 42,357 unique questions. When considering the union of questions and answers, 76,438 unique QA pairs were generated.

\begin{table}[htb]
\centering
\begin{tabular}{|c|c|c|}
\hline
\multirow{2}{*}{\textbf{Category}} & \multicolumn{2}{c|}{\textbf{Unique count}} \\  \cline{2-3}
 & \textbf{Questions} & \textbf{QA pairs} \\ \hline
Object recognition & 5,565 & 12,497 \\ 
Object state recognition & 4,729 & 7,627 \\ 
Object localization & 5,304 & 12,411 \\ 
Attribute recognition & 4,417 & 8,796 \\ 
Spatial reasoning & 8,968 & 12,364 \\ 
Functional reasoning & 5,656 & 10,831 \\ 
World knowledge & 7,718 & 11,912 \\ \hline
\textbf{Total} & \textbf{42,357} & \textbf{76,438 }\\ \hline
\end{tabular}
\caption{Counts of unique questions and QA pairs in the Ego4D VQA Gemini dataset}
\label{tab:qanda}
\end{table}

%% file: appendices/ego4d_gemini_examples.tex
\newpage
\section{Ego4D VQA Gemini Dataset Examples}
\label{sec:AppendixEgo4DVQAGeminiExamples}

\subsection{Domestic kitchen scene}

\begin{figure}[htbp]
\centering
\includegraphics[width=\linewidth]{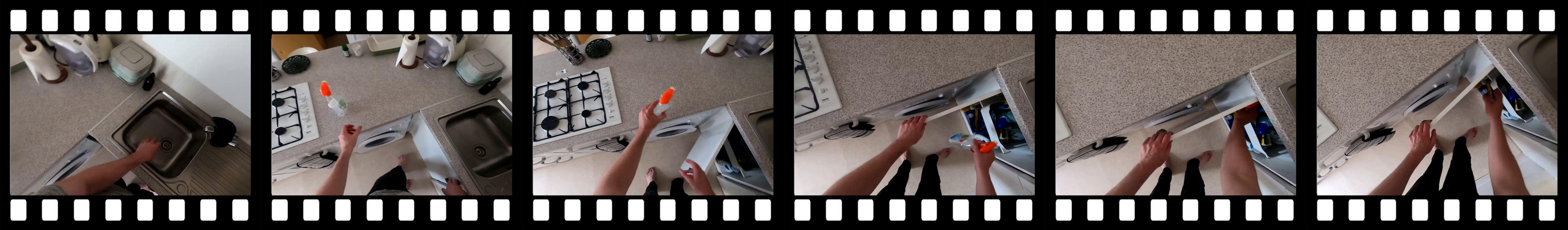}
\label{fig:frames_kitchen}
\end{figure}

\begin{tcolorbox}[width=\textwidth, colback=blue!10, colframe=black!100, title={Example of high quality generations in all categories}]
Category: Object recognition \\
Question: What household appliances are visible in the video?\\
Short answer: A washing machine and a stovetop \\\\
Category: Attribute recognition\\
Question: What color is the countertop?\\
Short answer: Gray\\\\
Category: Object state recognition\\
Question: Is the cabinet door open or closed at the start of the video?\\
Short answer: Closed\\\\
Category: Object localisation \\
Question: Where is the orange bottle placed?\\
Short answer: Inside the cabinet, to the left of the cleaning supplies\\\\
Category: Spatial reasoning\\
Question: Is the sink located to the left or right of the stovetop?\\
Short answer: Right \\\\
Category: Functional reasoning \\
Question: What is the purpose of the cabinet under the sink?\\
Short answer: To store cleaning supplies and other household items \\\\
Category: World knowledge\\
Question: What type of cleaner is typically used to clean a stovetop? \\
Short answer: Degreaser or multi-surface cleaner    
\end{tcolorbox}

\newpage
\subsection{Nighttime street scene}

\begin{figure}[htbp]
\centering
\includegraphics[width=\linewidth]{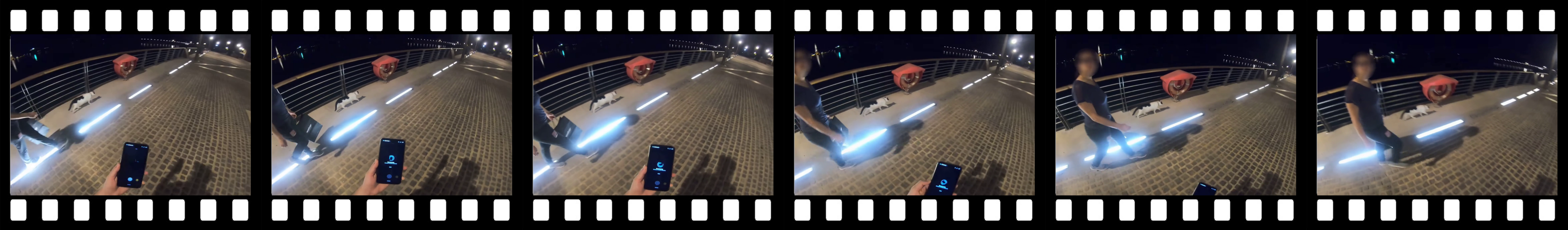}
\label{fig:frames_cat}
\end{figure}

\begin{tcolorbox}[width=\textwidth, colback=blue!10, colframe=black!100, title={Example of high quality generations in all but one category, with an incorrect answer for spatial reasoning}]
Category: object recognition\\
Question: What kind of animal is in the video? \\
Short answer: A cat\\
\\
Category: attribute recognition\\
Question: What color is the cat?\\
Short answer: Black and white \\
\\
Category: object state recognition\\
Question: Is the phone on or off? \\
Short answer: On \\
\\
Category: object localisation \\
Question: Where is the life preserver? \\
Short answer: Attached to the railing \\
\\
Category: spatial reasoning \\
Question: Is the cat closer to the person holding the phone or the person walking? \\
Short answer: The person holding the phone\\
\\
Category: functional reasoning\\
Question: What is the phone being used for? \\
Short answer: It is not clear what the phone is being used for \\
\\
Category: world knowledge\\
Question: What time of day is it?\\
Short answer: Night
\end{tcolorbox}

\newpage
\subsection{Shop counter scene}

\begin{figure}[htbp]
\centering
\includegraphics[width=\linewidth]{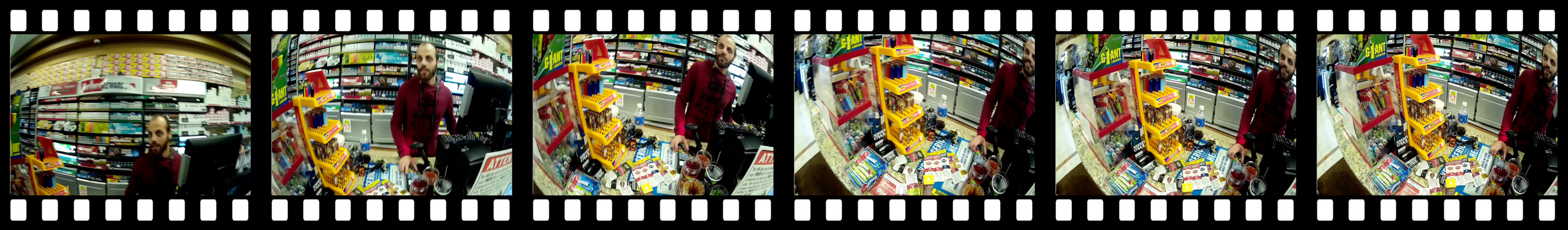}
\label{fig:frames_shop}
\end{figure}

\begin{tcolorbox}[width=\textwidth, colback=blue!10, colframe=black!100, title={Example of high quality generations in all but one category, with changing object state resulting in a partially correct answer for object state recognition}]
Category: object recognition \\
Question: What items are on display behind the cashier?\\
Short answer: cigarettes, cigars, chewing tobacco\\
\\
Category: attribute recognition \\
Question: What color is the cash register?\\
Short answer: black\\
\\
Category: object state recognition \\
Question: Is the cashier's mouth open or closed?\\
Short answer: open \\
\\
Category: object localisation\\
Question: Where is the display of lighters located?\\
Short answer: to the left of the cashier\\
\\
Category: spatial reasoning\\
Question: What is closer to the camera, the lighters or the cigarettes?\\
Short answer: the lighters\\
\\
Category: functional reasoning\\
Question: What is the purpose of the cash register?\\
Short answer: to process transactions\\
\\
Category: world knowledge\\
Question: What is the legal age to purchase tobacco products in the United States?\\
Short answer: 21 years old
\end{tcolorbox}

\newpage
\subsection{Outdoors dog scene}

\begin{figure}[htbp]
\centering
\includegraphics[width=\linewidth]{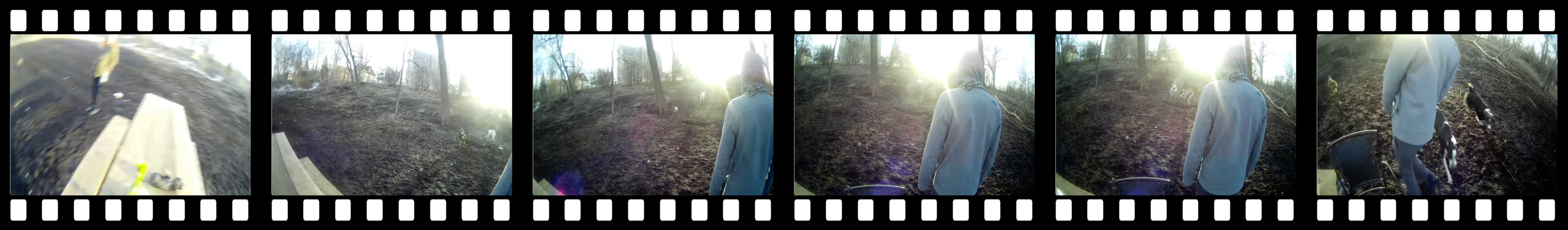}
\label{fig:frames_dogs}
\end{figure}

\begin{tcolorbox}[width=\textwidth, colback=blue!10, colframe=black!100, title={Example of hallucinated objects in questions, incorrect or over-confident answers for most categories}]
Category: Object recognition \\
Question: What breed of dog is in the video?\\
Short answer: There is a Bernese Mountain dog, a white Pitbull, and a black and white Border Collie.\\
\\
Category: Attribute recognition\\
Question: What color is the dog's collar?\\
Short answer: Red\\
\\
Category: Object state recognition\\
Question: Is the sun shining?\\
Short answer: Yes \\
\\
Category: Object localisation\\
Question: Where is the person standing? \\
Short answer: At the bottom of a hill\\
\\
Category: Spatial reasoning \\
Question: Are the dogs uphill or downhill from the person?\\
Short answer: Uphill\\
\\
Category: Functional reasoning\\
Question: What is the leash used for? \\
Short answer: To keep the dog from running away \\
\\
Category: World knowledge \\
Question: What season is it?\\
Short answer: It is most likely early spring
\end{tcolorbox}

\newpage
\subsection{Garden landscaping scene}

\begin{figure}[htbp]
\centering
\includegraphics[width=\linewidth]{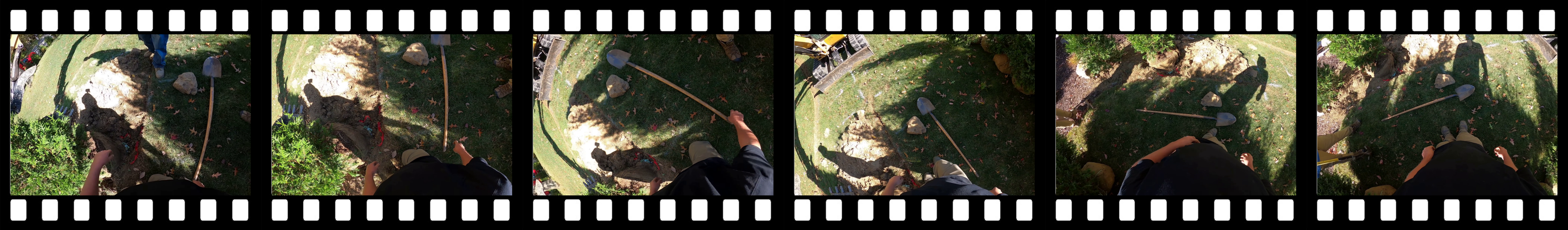}
\label{fig:frames_hole}
\end{figure}

\begin{tcolorbox}[width=\textwidth, colback=blue!10, colframe=black!100, title={Example of non-visually grounded answers for object recognition, object state recognition and functional reasoning categories}]
Category: object recognition\\
Question: What type of equipment is being used to dig the hole?\\
Short answer: A mini excavator.\\
\\
Category: attribute recognition\\
Question: What color is the handle of the shovel?\\
Short answer: Brown.\\
\\
Category: object state recognition\\
Question: Is the hole completely dug? \\
Short answer: No, it is still being dug. \\
\\
Category: object localisation\\
Question: Where is the red ribbon located?\\
Short answer:  Next to the hole that is being dug.\\
\\
Category: spatial reasoning\\
Question: Is the shovel closer to the mini excavator or the hole?\\
Short answer: The hole.\\
\\
Category: functional reasoning \\
Question: What is the purpose of the red ribbon? \\
Short answer: It is likely marking the location of underground utilities.\\
\\
Category: world knowledge\\
Question: What type of professional is likely operating the mini excavator?\\
Short answer: A landscaper or construction worker.
\end{tcolorbox}




\newpage
\subsection{Workshop scene}

\begin{figure}[htbp]
\centering
\includegraphics[width=\linewidth]{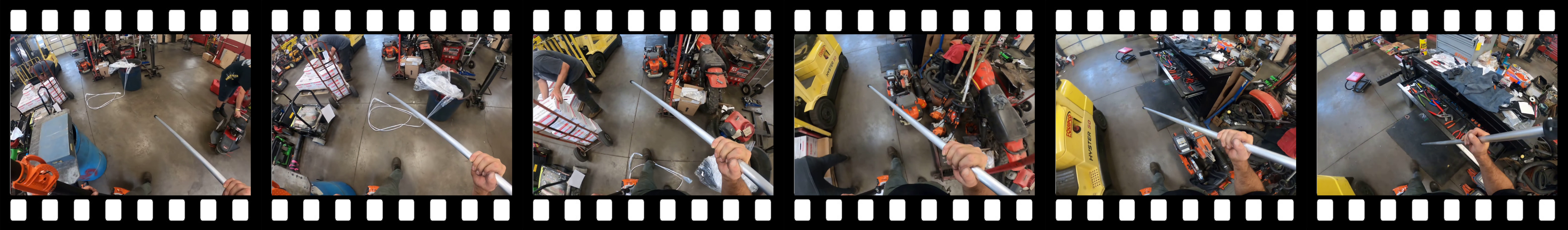}
\label{fig:frames_workshop}
\end{figure}

\begin{tcolorbox}[width=\textwidth, colback=blue!10, colframe=black!100, title={Example of hallucinated objects in questions and answers for object recognition, object state recognition, object localisation and spatial reasoning categories}]
Category: object recognition\\
Question: What types of equipment are visible in the workshop?\\
Short answer: The workshop contains various equipment, including chainsaws, leaf blowers, a forklift, hand tools, and a workbench.\\
\\
Category: attribute recognition \\
Question: What color is the forklift in the video?\\
Short answer: Yellow\\
\\
Category: object state recognition \\
Question: Are the chainsaws in the video new or used? \\
Short answer: It is difficult to tell from the video alone whether the chainsaws are new or used. \\
\\
Category: object localisation \\
Question: Where is the workbench located? \\
Short answer: The workbench is located on the right side of the video, against the wall.\\
\\
Category: spatial reasoning \\
Question: How many chainsaws are to the left of the red toolbox? \\
Short answer: There are five chainsaws to the left of the red toolbox. \\
\\
Category: functional reasoning\\
Question: What is the purpose of the extension pole shown in the video?\\
Short answer: The extension pole is likely used to reach high places, possibly for tasks like pruning trees or cleaning gutters. \\
\\
Category: world knowledge \\
Question: What types of businesses typically use forklifts for their operations? \\
Short answer:  Forklifts are commonly used in warehouses, distribution centers, construction sites, and other businesses that handle heavy materials.
\end{tcolorbox}

\newpage
\normalsize

%% file: appendices/ego4d_gemini_evaluation.tex
\section{Human Evaluation of Gemini Pro 1.5 Generated Training Data}
\label{sec:AppendixGeminiEvaluation}

Following the human evaluation schema described for the Self-Instruct dataset \citep{Self-Instruct}, an author of the present work evaluated the 1,400 examples as follows:
\begin{enumerate}
    \item Is the question appropriate for the clip?
    \item Is the question appropriate for the category?
    \item Is the answer correct and acceptable for the clip and question?
\end{enumerate}

\subsection{Results}

\begin{figure}[htbp]
\centering
\includegraphics[width=0.6\linewidth]{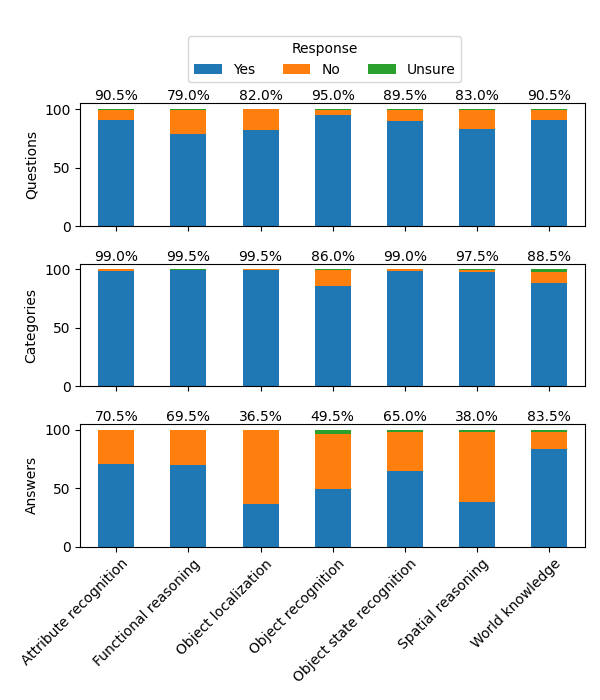}
\caption{Results of human evaluation on 1,400 examples. The percentage of appropriate question, appropriate category, and correct answer are shown on a per category basis. Text labels show the percentage of questions/categories/answers in each category found to be appropriate and/or correct.}
\label{fig:human_eval_cqa}
\end{figure}

Results of the human evaluation are shown in \cref{fig:human_eval_cqa}. The rates of appropriate generated questions ranged from 79.0\% for the functional reasoning category to 95.0\% for the object recognition category. For assigning the questions to appropriate categories, Gemini performed favourably, with a range of 86.0\% for object recognition to 99.5\% for functional reasoning and object localization. Gemini performed markedly worse with generating correct and acceptable answers for the clips, ranging from 36.5\% for object localization to 83.5\% for world knowledge. The superior performance in the world knowledge category could be due to the advantages of relying on the language model's encoded knowledge, without the need to refer to the visual context of the scene. In addition, VLMs have an observed concept association bias and weakness in compositional understanding, with tasks such as spatial reasoning being especially prone to errors (\citealp{yamada2022}; \citealp{yuksekgonul2022}).

Issues with the generated questions and answers often included hallucinated objects, non-visually grounded answers and changing camera angles resulting in partially correct answers. See \cref{sec:AppendixEgo4DVQAGeminiExamples} for examples of generated \textit{(category, question, answer)} tuples.

\subsection{Inter-Annotator Agreement}

In addition, 10\% of the 200 clips were randomly chosen for evaluation by another human expert and used to estimate inter-annotator agreement by calculating Cohen's Kappa. Agreement between annotators was found to be fair for question and category appropriateness, with scores of 0.210 and 0.232 respectively. For answer correctness and acceptability, agreement was found to be moderate, with a score of 0.403. Although the scores show an agreement between experts, they also indicate the difficulty of evaluating generated questions and answers, with clips often changing state, for example, via object movement and multiple camera angles resulting in changing spatial relationships.

%% file: appendices/vsr_prompt.tex
\section{VSR Prompt}
\label{sec:AppendixVsrPrompt}

\begin{tcolorbox}[width=\textwidth, colback=gray!10, colframe=black!100, title=VSR prompt]
Generate a polar question from the following statement about a picture.
Keep as many words as you can of the statement in the question and do not add unnecessary words.
Always generate just the question.
Do not include any explanations.

Statement: \texttt{<statement>}
\end{tcolorbox}

%% file: appendices/egoclip.tex
\section{EgoClip Preprocessing}
\label{sec:AppendixEgoclipPreprocessing}

Given the original EgoClip dataset, we preprocess it using specific rules to convert it into a more natural caption. Specifically, we first sample a prompt from a list of predefined prompts (see below) and then apply conversion rules to the original caption. Specifically, following the Ego4D guidelines~\footnote{\url{https://ego4d-data.org/docs/data/annotation-guidelines/\#narrations}}, we replace ``\#C" with ``the camera wearer", ``\#O" with ``another person", and ``\#UNSURE" with ``something". Finally, we delete the prefix ``Summary" when included.    

\begin{tcolorbox}[width=\textwidth, colback=gray!10, colframe=black!100, title=EgoClip prompt instructions]

  [ \\
        "Can you please provide a brief description of the video?", \\
        "Describe the content of the video.", \\
        "What is happening in the video? Please describe it.", \\
        "Can you summarize the key events or actions in the video?", \\
        "Describe the visual elements and any notable features in the video.", \\
        "Provide a narrative description of the video.", \\
        "What's in the video?", \\
        "What can you see in this video?", \\
        "What's happening in the video?", \\
        "What is the main focus of the video?" \\
    ]
\end{tcolorbox}

%% file: appendices/training.tex
\section{Training Details}
\label{sec:AppendixTrainingDetails}
We build our rehearsal dataset composed of previously learned examples starting from the Chat-UniVi instruction tuning dataset, which includes instances from LLaVa (\citealp{liu2024-llava}; composed of NLP and COCO examples), MIMIC-IT~\cite{li2023mimic}, and Video-ChatGPT \cite{maaz2023-videochatgpt}. Since we wanted \model to forget language skills as little as possible, and to have good video understanding capabilities, we adapted the distribution of previously learned examples, giving slightly less emphasis to the text and image instances, and much more emphasis to the Video-ChatGPT instances. In particular, we bring the percentage of LLaVa NLP instances, LLaVa COCO instances, MIMIC-IT instances, and Video-ChatGPT from 5\%, 82\%, 13\%, and 25\% to 10\%, 20\%, 50\%, and 20\%. Given the instances resulting from the changed distribution, we sample 1\% from each subset in order to build the rehearsal data leveraged in our experiments which is composed of 10,000 instances (see \cref{tab:rehearsal_data}).

\begin{table}
\centering
\begin{tabular}{|l|c|c|}
\hline
\textbf{Subset} & \multicolumn{1}{l|}{\textbf{Sampled instances}} & \multicolumn{1}{l|}{\textbf{Sampling percentage}} \\ \hline
NLP             & 1000                                            & 10      \\                              
COCO            & 2000                                            & 20        \\                           
VideoChat       & 5000                                            & 50      \\                             
MIMIC           & 2000                                            & 20                                         \\ \hline
\textbf{Total}           & \textbf{10000}                                           & \textbf{100}                                               \\ \hline
\end{tabular}
\caption{Proportion of data used for vision+language rehearsal during our fine-tuning stage. Data are derived from several data sources used for Chat-UniVi instruction-tuning~\cite{jin2023chat}.}
\label{tab:rehearsal_data}
\end{table}

Following best practices in using LoRa\footnote{ \url{https://lightning.ai/pages/community/lora-insights/}}, we employ the Adam optimizer with a learning rate equal to 3e-4 to fine-tune for one epoch and we set the rank $R$ equal to 64 and value of $\alpha$ equal to 128.

\subsection{Computational Experiments}

\model is a 7B parameter model trained using A10 NVIDIA GPUs available in AWS. Each training run lasted approximately 8 hours on a single GPU thanks to LoRA. Running all the configurations of \model required an overall computational budget of 80 GPU/hours. 

%% file: appendices/results.tex
\section{Extended Evaluation \& Results}
\label{sec:AppendixResults}

\subsection{Response Generation}

\subsubsection{ChatUniVi Variants}

To generate the ChatUniVi responses, we use the default parameters, i.e., we set the temperature of the model to 0.2 and use beam search with a single beam. For processing the input videos, we consider two approaches. For the first approach, we process the videos by sampling frames at a rate of 1 frame per second. Then, if there are more than a maximum of 100 frames, we resample 100 frames uniformly from the sampled set. For the second approach, we sample 50 frames uniformly at the original frame rate of the video, if there are more than 50 frames. Otherwise, we use all the available frames. Frames are also resized to $224 \times 224$ pixels, as per the original model resolution~\cite{jin2023chat}.

\subsubsection{Gemini 1.5 Variants}

We also evaluate two variants of Gemini 1.5, i.e., the Pro and Flash variants. For this, we used a similar protocol as the one used for the evaluation of Gemini 1.0 Pro Vision on the OpenEQA benchmark~\cite{OpenEQA2023}. We accessed these models through the Vertex AI API\footnote{\url{https://cloud.google.com/vertex-ai}}. The prompt was constructed by concatenating the prompt and the frames as follows:

\begin{tcolorbox}[width=\textwidth, colback=gray!10, colframe=black!100, title=Gemini prompt]
You are an intelligent question answering agent. I will ask you questions about an indoor space and you must provide an answer.\\
You will be shown a set of images that have been collected from a single location.\\
Given a user query, you must output `text` to answer to the question asked by the user.\\
<FRAME> ... <FRAME>\\
User Query: {question}
\end{tcolorbox}

By submitting the frames instead of the video, we could control the number of frames that were sent to the model. In the results section, we show two sets of results. One set of results used the full frame sizes and the other set corresponds to sending frames resized to $224 \times 224$ pixels.

\subsection{Evaluation Protocol}

To evaluate the models in this paper, we follow an evaluation protocol inspired by the OpenEQA benchmark~\cite{OpenEQA2023}. Concretely, we submit the prompt below to a Llama3 70B model~\cite{llama3modelcard} through the together.ai API\footnote{\url{https://www.together.ai/}}. There is another variant for examples that include "extra answers", which follows a similar format, but the model is also prompted to check the extra answers to make an assessment as to whether the generated response answers the given question. As illustrated in the prompt, the Llama3 model is prompted to give a score between 1 and 5 depending on how well the generated response matches any of the ground-truth answers. Once we have obtained the scores for all the samples in the dataset, we normalise them and compute their mean and bootstrapped standard error.

\begin{tcolorbox}[width=\textwidth, colback=gray!10, colframe=black!100, title=Llama3 prompt]
You are an AI assistant who will help me to evaluate the response given the question and the correct answer.\\
To mark a response, you should output a single integer between 1 and 5 (including 1, 5).\\
5 means that the response perfectly matches the answer.\\
1 means that the response is completely different from the answer.\\

Example 1:\\
Question: Is it overcast?\\
Answer: no\\
Response: yes\\
Your mark: 1\\

Example 2:\\
Question: Who is standing at the table?\\
Answer: woman\\
Response: Jessica\\
Your mark: 3\\

Example 3:\\
Question: Are there drapes to the right of the bed?\\
Answer: yes\\
Response: yes\\
Your mark: 5\\

Your Turn:\\
Question: {question}\\
Answer: {answer}\\
Response: {prediction}
\end{tcolorbox}


\subsection{Results}

Table \ref{table:AppendixDetailedResults} shows the results per category, per subset, and for all instances of blind models, VLMs, and AlanaVLM's ablations. When it comes to ablations, we evaluated different mixtures of the \dataset to verify the impact of different data sources on the overall performance in the OpenEQA benchmark. Additionally, we also experimented with different numbers of video frames. 
Following the OpenEQA evaluation protocol, we use bootstrapping to estimate standard deviations associated with the different model configurations.

\begin{sidewaystable*}[!ht]
\footnotesize
\centering
\begin{tabular}{|lrrrrrrrrrr|}
\hline
\multicolumn{1}{|l|}{\textbf{Model}} & \multicolumn{1}{c|}{\textbf{\begin{tabular}[c]{@{}c@{}}Object\\ Recognition\end{tabular}}} & \multicolumn{1}{c|}{\textbf{\begin{tabular}[c]{@{}c@{}}Object State\\ Recognition\end{tabular}}} & \multicolumn{1}{c|}{\textbf{\begin{tabular}[c]{@{}c@{}}Object\\ Localisation\end{tabular}}} & \multicolumn{1}{c|}{\textbf{\begin{tabular}[c]{@{}c@{}}Attribute\\ Recognition\end{tabular}}} & \multicolumn{1}{c|}{\textbf{\begin{tabular}[c]{@{}c@{}}Spatial\\ Understanding\end{tabular}}} & \multicolumn{1}{c|}{\textbf{\begin{tabular}[c]{@{}c@{}}Functional\\ Reasoning\end{tabular}}} & \multicolumn{1}{c|}{\textbf{\begin{tabular}[c]{@{}c@{}}World\\ Knowledge\end{tabular}}} & \multicolumn{1}{c|}{\textbf{SN}} & \multicolumn{1}{c|}{\textbf{HM3D}} & \multicolumn{1}{c|}{\textbf{All}} \\ \hline
\multicolumn{11}{|l|}{\textbf{Blind LLMs}} \\ \hline
\multicolumn{1}{|l|}{GPT-4*} & \multicolumn{1}{r|}{15.4} & \multicolumn{1}{r|}{51} & \multicolumn{1}{r|}{20.3} & \multicolumn{1}{r|}{31.5} & \multicolumn{1}{r|}{31.4} & \multicolumn{1}{r|}{52.2} & \multicolumn{1}{r|}{34.2} & \multicolumn{1}{r|}{32.5 ± 1.2} & \multicolumn{1}{r|}{35.5 ± 1.7} & 33.5 ± 1.0 \\
\multicolumn{1}{|l|}{Chat-UniVi (text-only)} & \multicolumn{1}{r|}{33.1 ± 2.6} & \multicolumn{1}{r|}{55.5 ± 3.0} & \multicolumn{1}{r|}{24.0 ± 2.4} & \multicolumn{1}{r|}{29.2 ± 2.8} & \multicolumn{1}{r|}{38.2 ± 2.9} & \multicolumn{1}{r|}{51.4 ± 2.7} & \multicolumn{1}{r|}{48.9 ± 2.9} & \multicolumn{1}{r|}{43.4 ± 1.3} & \multicolumn{1}{r|}{32.4 ± 1.7} & 39.7 ± 1.1 \\ \hline
\multicolumn{11}{|l|}{\textbf{Proprietary Multi-Frame VLMs}} \\ \hline
\multicolumn{1}{|l|}{GPT-4V (50f)*} & \multicolumn{1}{r|}{51.4} & \multicolumn{1}{r|}{57.7} & \multicolumn{1}{r|}{53.3} & \multicolumn{1}{r|}{65.2} & \multicolumn{1}{r|}{42.6} & \multicolumn{1}{r|}{63.8} & \multicolumn{1}{r|}{52.3} & \multicolumn{1}{r|}{57.4 ± 1.3} & \multicolumn{1}{r|}{51.3 ± 1.8} & 55.3 ± 1.1 \\
\multicolumn{1}{|l|}{Claude 3 (20f)*} & \multicolumn{1}{r|}{37.0} & \multicolumn{1}{r|}{45.5} & \multicolumn{1}{r|}{13.1} & \multicolumn{1}{r|}{39.2} & \multicolumn{1}{r|}{37.0} & \multicolumn{1}{r|}{37.9} & \multicolumn{1}{r|}{47.3} & \multicolumn{1}{r|}{n/a} & \multicolumn{1}{r|}{n/a} & 36.3 ± 1.1 \\
\multicolumn{1}{|l|}{Gemini 1.0 Pro V. (15f)*} & \multicolumn{1}{r|}{41.5} & \multicolumn{1}{r|}{56.9} & \multicolumn{1}{r|}{33.3} & \multicolumn{1}{r|}{41.9} & \multicolumn{1}{r|}{37.6} & \multicolumn{1}{r|}{52.2} & \multicolumn{1}{r|}{52.1} & \multicolumn{1}{r|}{n/a} & \multicolumn{1}{r|}{n/a} & 44.9 ± 1.1 \\
\multicolumn{1}{|l|}{Gemini 1.5 Flash (50f)} & \multicolumn{1}{r|}{\textbf{73.6} ± 2.6} & \multicolumn{1}{r|}{\textbf{76.0} ± 2.6} & \multicolumn{1}{r|}{61.4 ± 2.3} & \multicolumn{1}{r|}{\textbf{81.8} ± 2.2} & \multicolumn{1}{r|}{56.7 ± 3.0} & \multicolumn{1}{r|}{\textbf{78.3} ± 2.2} & \multicolumn{1}{r|}{\textbf{81.1} ± 2.2} & \multicolumn{1}{r|}{\textbf{74.0} ± 1.1} & \multicolumn{1}{r|}{\textbf{69.7} ± 1.7} & \multicolumn{1}{r|}{\textbf{72.5} ± 0.9} \\
\multicolumn{1}{|l|}{Gemini 1.5 Flash (50f - 224 x 224)} & \multicolumn{1}{r|}{71.0 ± 2.7} & \multicolumn{1}{r|}{75.5 ± 2.6} & \multicolumn{1}{r|}{\textbf{62.8} ± 2.4} & \multicolumn{1}{r|}{80.8 ± 2.2} & \multicolumn{1}{r|}{55.9 ± 3.0} & \multicolumn{1}{r|}{76.8 ± 2.2} & \multicolumn{1}{r|}{74.1 ± 2.6} & \multicolumn{1}{r|}{71.9 ± 1.2} & \multicolumn{1}{r|}{69.1 ± 1.7} & \multicolumn{1}{r|}{71.0 ± 1.0} \\
\multicolumn{1}{|l|}{Gemini 1.5 Pro (50f)} & \multicolumn{1}{r|}{73.1 ± 2.6} & \multicolumn{1}{r|}{60.9 ± 2.7} & \multicolumn{1}{r|}{56.3 ± 2.5} & \multicolumn{1}{r|}{74.4 ± 2.4} & \multicolumn{1}{r|}{\textbf{59.4} ± 3.0} & \multicolumn{1}{r|}{63.5 ± 2.7} & \multicolumn{1}{r|}{67.6 ± 2.7} & \multicolumn{1}{r|}{66.9 ± 1.2} & \multicolumn{1}{r|}{61.0 ± 1.8} & \multicolumn{1}{r|}{64.9 ± 1.0} \\
\multicolumn{1}{|l|}{Gemini 1.5 Pro (50f - 224 x 224)} & \multicolumn{1}{r|}{69.0 ± 2.7} & \multicolumn{1}{r|}{61.4 ± 2.7} & \multicolumn{1}{r|}{53.0 ± 2.6} & \multicolumn{1}{r|}{69.7 ± 2.5} & \multicolumn{1}{r|}{55.6 ± 3.0} & \multicolumn{1}{r|}{61.5 ± 2.7} & \multicolumn{1}{r|}{63.5 ± 2.8} & \multicolumn{1}{r|}{64.3 ± 1.3} & \multicolumn{1}{r|}{57.1 ± 1.8} & \multicolumn{1}{r|}{61.9 ± 1.0} \\ \hline
\multicolumn{11}{|l|}{\textbf{Open-Source Multi-Frame VLMs}} \\ \hline
\multicolumn{1}{|l|}{Chat-UniVi} & \multicolumn{1}{r|}{28.9 ± 2.6} & \multicolumn{1}{r|}{57.1 ± 3.0} & \multicolumn{1}{r|}{23.8 ± 2.4} & \multicolumn{1}{r|}{35.6 ± 2.9} & \multicolumn{1}{r|}{37.4 ± 2.9} & \multicolumn{1}{r|}{59.1 ± 2.7} & \multicolumn{1}{r|}{52.7 ± 2.9} & \multicolumn{1}{r|}{42.6 ± 1.3} & \multicolumn{1}{r|}{39.8 ± 1.9} & 41.7 ± 1.1 \\
\multicolumn{1}{|l|}{Chat-UniVi (50f)} & \multicolumn{1}{r|}{33.8 ± 2.7} & \multicolumn{1}{r|}{45.1 ± 2.8} & \multicolumn{1}{r|}{27.9 ± 2.5} & \multicolumn{1}{r|}{33.4 ± 2.9} & \multicolumn{1}{r|}{44.5 ± 3.0} & \multicolumn{1}{r|}{\textbf{63.8} ± 2.6} & \multicolumn{1}{r|}{52.0 ± 3.0} & \multicolumn{1}{r|}{43.4 ± 1.3} & \multicolumn{1}{r|}{40.4 ± 1.8} & 42.3 ± 1.1 \\
\multicolumn{1}{|l|}{Chat-UniVi (Rehearsal)} & \multicolumn{1}{r|}{32.3 ± 2.6} & \multicolumn{1}{r|}{55.5 ± 3.0} & \multicolumn{1}{r|}{26.8 ± 2.4} & \multicolumn{1}{r|}{38.0 ± 3.0} & \multicolumn{1}{r|}{43.3 ± 3.0} & \multicolumn{1}{r|}{57.5 ± 2.7} & \multicolumn{1}{r|}{\textbf{58.3} ± 2.9} & \multicolumn{1}{r|}{45.7 ± 1.4} & \multicolumn{1}{r|}{40.8 ± 1.9} & 44.0 ± 1.1 \\
\multicolumn{1}{|l|}{Chat-UniVi (Rehearsal) (50f)} & \multicolumn{1}{r|}{36.1 ± 2.7} & \multicolumn{1}{r|}{44.5 ± 2.7} & \multicolumn{1}{r|}{27.9 ± 2.4} & \multicolumn{1}{r|}{35.6 ± 2.9} & \multicolumn{1}{r|}{44.4 ± 3.0} & \multicolumn{1}{r|}{57.1 ± 2.8} & \multicolumn{1}{r|}{52.2 ± 2.9} & \multicolumn{1}{r|}{42.9 ± 1.3} & \multicolumn{1}{r|}{40.3 ± 1.8} & 42.0 ± 1.1 \\
\multicolumn{1}{|l|}{AlanaVLM (VQA-EgoClip)} & \multicolumn{1}{r|}{30.1 ± 2.5} & \multicolumn{1}{r|}{56.2 ± 3.1} & \multicolumn{1}{r|}{29.0 ± 2.4} & \multicolumn{1}{r|}{41.7 ± 3.0} & \multicolumn{1}{r|}{\textbf{45.7} ± 3.0} & \multicolumn{1}{r|}{61.1 ± 2.5} & \multicolumn{1}{r|}{51.9 ± 3.0} & \multicolumn{1}{r|}{47.0 ± 1.4} & \multicolumn{1}{r|}{40.2 ± 1.9} & 44.7 ± 1.1 \\
\multicolumn{1}{|l|}{AlanaVLM (VQA-EgoClip) (50f)} & \multicolumn{1}{r|}{\textbf{39.8} ± 2.7} & \multicolumn{1}{r|}{54.9 ± 3.1} & \multicolumn{1}{r|}{\underline{32.0} ± 2.4} & \multicolumn{1}{r|}{42.7 ± 3.0} & \multicolumn{1}{r|}{\underline{45.5} ± 3.0} & \multicolumn{1}{r|}{59.6 ± 2.5} & \multicolumn{1}{r|}{52.5 ± 2.9} & \multicolumn{1}{r|}{\underline{47.4} ± 1.3} & \multicolumn{1}{r|}{\underline{44.3} ± 1.9} & \underline{46.3} ± 1.1 \\
\multicolumn{1}{|l|}{AlanaVLM (VQA-EgoClip-HM3D)} & \multicolumn{1}{r|}{33.2 ± 2.7} & \multicolumn{1}{r|}{56.3 ± 3.1} & \multicolumn{1}{r|}{31.2 ± 2.5} & \multicolumn{1}{r|}{40.2 ± 3.0} & \multicolumn{1}{r|}{41.7 ± 3.0} & \multicolumn{1}{r|}{61.2 ± 2.5} & \multicolumn{1}{r|}{53.6 ± 3.0} & \multicolumn{1}{r|}{46.8 ± 1.3} & \multicolumn{1}{r|}{41.5 ± 1.9} & 45.0 ± 1.1 \\
\multicolumn{1}{|l|}{AlanaVLM (VQA-EgoClip-HM3D) (50f)} & \multicolumn{1}{r|}{36.4 ± 2.7} & \multicolumn{1}{r|}{53.9 ± 3.1} & \multicolumn{1}{r|}{30.5 ± 2.4} & \multicolumn{1}{r|}{\underline{44.5} ± 3.1} & \multicolumn{1}{r|}{38.0 ± 2.9} & \multicolumn{1}{r|}{56.8 ± 2.6} & \multicolumn{1}{r|}{\underline{56.1} ± 3.0} & \multicolumn{1}{r|}{45.9 ± 1.3} & \multicolumn{1}{r|}{42.7 ± 1.8} & 44.8 ± 1.1 \\
\multicolumn{1}{|l|}{AlanaVLM (VQA-VSR-EgoClip) (50f)} & \multicolumn{1}{r|}{32.0 ± 2.6} & \multicolumn{1}{r|}{50.5 ± 3.1} & \multicolumn{1}{r|}{29.3 ± 2.5} & \multicolumn{1}{r|}{41.8 ± 3.0} & \multicolumn{1}{r|}{42.7 ± 3.0} & \multicolumn{1}{r|}{61.1 ± 2.6} & \multicolumn{1}{r|}{50.7 ± 3.0} & \multicolumn{1}{r|}{45.4 ± 1.4} & \multicolumn{1}{r|}{40.0 ± 1.9} & 43.6 ± 1.1 \\
\multicolumn{1}{|l|}{AlanaVLM (VQA-VSR-EgoClip) (50f)} & \multicolumn{1}{r|}{\underline{37.1} ± 2.6} & \multicolumn{1}{r|}{\underline{57.5} ± 3.1} & \multicolumn{1}{r|}{31.0 ± 2.5} & \multicolumn{1}{r|}{\textbf{46.2} ± 3.1} & \multicolumn{1}{r|}{43.4 ± 3.0} & \multicolumn{1}{r|}{\underline{61.9} ± 2.5} & \multicolumn{1}{r|}{52.5 ± 3.0} & \multicolumn{1}{r|}{\textbf{47.8} ± 1.4} & \multicolumn{1}{r|}{\textbf{44.8} ± 1.9} & \textbf{46.7} ± 1.1 \\
\multicolumn{1}{|l|}{AlanaVLM (VQA-VSR-EgoClip-HM3D)} & \multicolumn{1}{r|}{32.7 ± 2.7} & \multicolumn{1}{r|}{\textbf{59.4} ± 3.0} & \multicolumn{1}{r|}{\textbf{36.6} ± 2.6} & \multicolumn{1}{r|}{39.2 ± 3.0} & \multicolumn{1}{r|}{37.2 ± 2.9} & \multicolumn{1}{r|}{\underline{61.9} ± 2.6} & \multicolumn{1}{r|}{54.1 ± 3.0} & \multicolumn{1}{r|}{47.2 ± 1.4} & \multicolumn{1}{r|}{42.6 ± 1.9} & 45.6 ± 1.1 \\
\multicolumn{1}{|l|}{AlanaVLM (VQA-VSR-EgoClip-HM3D) (50f)} & \multicolumn{1}{r|}{37.0 ± 2.6} & \multicolumn{1}{r|}{55.4 ± 3.1} & \multicolumn{1}{r|}{30.7 ± 2.5} & \multicolumn{1}{r|}{43.9 ± 3.1} & \multicolumn{1}{r|}{40.5 ± 2.9} & \multicolumn{1}{r|}{58.6 ± 2.5} & \multicolumn{1}{r|}{50.2 ± 2.9} & \multicolumn{1}{r|}{46.7 ± 1.3} & \multicolumn{1}{r|}{41.4 ± 1.8} & 44.9 ± 1.1 \\ \hline
\end{tabular}
\caption{Results per category, per subset, and for all instances of blind models, VLMs, and AlanaVLM ablations (with \textit{n}f indicating the number of frames). Standard deviations were estimated using bootstrapping as per the OpenEQA evaluation protocol \citep{OpenEQA2023}. In this table, we refer to the union of Ego4D VQA NLQ human annotated QA pairs (\cref{sec:ego4d_vqa}) and Ego4D VQA Gemini (\cref{sec:ego4d_vqa_gemini}) as VQA. (*): Results taken from \citet{OpenEQA2023}.}
\label{table:AppendixDetailedResults}
\end{sidewaystable*}

%% file: appendices/alanavlm_evaluation.tex
\section{Error Analysis of AlanaVLM's performance}
\label{sec:AppendixAlanaVLMEvaluation}

The subset of examples used for the human evaluation of AlanaVLM's performance has been obtained through stratified sampling based on question categories for each dataset.  Since we have seven question types per category and two subsets (ScanNet and HM3D), we obtained 98 examples. The mistakes made by AlanaVLM were pointed out by two authors of the present work who provided ratings and categorised the errors according to special categories that were created in a bottom-up fashion. To compute the percentage of correct answers according to humans, we counted the number of times where human ratings were $\ge 4$ and LLM ratings were $\le 2$.

%% file: appendices/data_model_release.tex
\section{Data and Model Release Details}

We will release both the EVUD dataset as well as the trained checkpoints that were produced in the context of this paper alongside their predictions for the OpenEQA benchmark. We plan to release the model checkpoints and code under MIT license. On the other hand, we will release the EVUD under CC BY 4.0. All these artefacts will be released on Huggingface Hub upon acceptance.